%% file: main.tex
\documentclass[11pt]{article}
\usepackage[margin=1in]{geometry}
\usepackage{amsmath,amssymb}
\usepackage{booktabs}
\usepackage{graphicx}
\usepackage{float}
\usepackage{algorithm}
\usepackage{algpseudocode}
\usepackage{hyperref}
\usepackage{natbib}

\title{Topology-Preserving Multi-Usage Pipe Centerline Refinement\\
under Permanent Tubular Occupancy}
\author{Deng Quan}
\date{}

\begin{document}
\maketitle

\input{sec_abstract}
\input{sec_intro}
\input{sec_problem}
\input{sec_related}
\input{sec_method}
\input{sec_costs}
\input{sec_experiments}

\input{sec_conclusion}

\bibliographystyle{plainnat}
\bibliography{refs}

\end{document}

%% file: sec_abstract.tex
% !TEX root = main.tex
% =============================================================================
% Abstract
% =============================================================================

\begin{abstract}
Discrete multi-group pipe routers place channels on a grid or coarse graph.
The resulting centerlines are topologically feasible but typically contain many
axis-aligned corners. For permanently occupied fluid tubes---as in
additively manufactured hydraulic manifolds---those corners are undesirable:
they degrade geometric fairness and, in practice, flow quality.
This technical report studies the downstream geometric stage that starts from a
fixed routing topology and a feasible coarse layout, and refines tubular
centerlines under permanent clearance.

Unlike classical robotic path smoothing, which usually fair one temporary
open trajectory, our setting is a branched manifold with declared terminal
usages, finite tube radii, and static cross-group occupancy.
We formalize the refinement problem, highlight the multi-usage junction
conflict that a single shared branch point cannot resolve, and outline a
hybrid trunk-share / junction-split parameterization with soft geometric costs
and hard clearance acceptance.
On a multi-group suite ($8$ groups $\times$ $8$ terminals, $r{=}2$) with
capsule clearance, the refine reaches full hard-accept success,
$\kappa r\lesssim 1$, and about a $95\%$ relative drop in the soft objective.
Physics-based pressure-drop optimization is intentionally deferred to a later
PDE stage; here the objective remains geometric.
\end{abstract}

\paragraph{Code.}
Reference implementation, instance loaders, and batch runners are in
\url{https://github.com/bladesaber/MAPF_Pipeline}
(branch \texttt{branch\_quan}), under \texttt{pipe\_smooth/}.

%% file: sec_intro.tex
% !TEX root = main.tex
% =============================================================================
% Introduction / background
% =============================================================================

\section{Introduction}
\label{sec:intro}

Additive manufacturing (AM) allows compact hydraulic components whose internal
channels follow free-form three-dimensional paths rather than intersecting
drilled holes.
A representative design case is multi-group channel layout in rotary
direct-drive servo valves: terminals are partitioned into groups; pipes of
different groups must remain geometrically separated for all poses of interest,
while pipes within a group may share trunks and branch into tree-like
manifolds.
Once a channel is placed it permanently occupies volume, so later groups must
clear earlier geometry.

Our companion work on discrete multi-group pipe routing under permanent
geometric occupancy~\cite{deng2025pipe_design} addresses the combinatorial
layout problem: given ports, obstacles, and group topologies, produce a
conflict-free coarse embedding on a constructive 3D grid benchmark, evaluated
by fixed-budget success of CBS, PBS, and priority planning.
That paper explicitly leaves post-layout smoothing as orthogonal follow-on
work.
The layouts it returns are feasible but still grid-angular; the present report
refines those centerlines.

\paragraph{Why robotic path smoothing is not enough.}
Trajectory fairing methods such as CHOMP, STOMP, TEB, and TrajOpt typically
optimize a single open curve for a temporary agent under time-indexed or soft
obstacle costs~\cite{zucker2013chomp,kalakrishnan2011stomp,roesmann2013teb,schulman2014trajopt}.
Pipe refinement differs along several axes that we make precise in
Sec.~\ref{sec:problem}: permanent tubular occupancy, branched topology with
degree-$\ge 3$ junctions, costs driven by declared usages rather than one
$s$--$t$ path.
The central geometric conflict is not ``make one spline smoother,'' but
\emph{how several overlapping usages share a trunk while remaining fair at every
declared terminal-to-terminal flow}.

\paragraph{Scope of this report.}
We formalize topology-preserving, multi-usage geometric refinement
under permanent tubular occupancy (Sec.~\ref{sec:problem}), survey adjacent
literatures (Sec.~\ref{sec:related}), and describe a hybrid trunk-share /
junction-split Hermite parameterization (Sec.~\ref{sec:method}) with soft path
and clearance costs (Sec.~\ref{sec:costs}).
Gradient-based experiments on synthetic instances with capsule clearance follow
in later sections.
We do not claim a new optimal continuous planner, commercial CAD parity, or
CFD-validated hydraulic improvement; those belong to later CAD and fluid
optimization, not to this geometric pass.

\paragraph{Contributions.}
\begin{enumerate}
  \item \textbf{Problem.} A clear formulation of post-routing centerline
        refinement for branched, multi-usage, permanently occupied tubes,
        including the shared-junction versus locally split junction regimes.
  \item \textbf{Method sketch.} Hybrid trunk-share / junction-split variables
        with soft length/curvature and clearance costs, hard ports and
        topology by construction, and hard clearance at accept
        (detailed in later sections).
  \item \textbf{Empirical note.} On hard multi-group synthetic cases, gradient
        refine with capsule clearance yields feasible designs with
        $\kappa r_{\max}\lesssim 1$ and large relative objective drop
        (Sec.~\ref{sec:experiments}).
\end{enumerate}

%% file: sec_problem.tex
% !TEX root = main.tex
% =============================================================================
% Problem formulation
% =============================================================================

\section{Problem Formulation}
\label{sec:problem}

\subsection{Setting}
\label{sec:problem:setting}

The geometric refine starts after a discrete multi-group router has returned a
feasible coarse layout.
We treat the following as \emph{fixed}:

\begin{itemize}
  \item \textbf{Domain} $\Omega$: a bounded design region with static obstacles
        and already committed pipes of other groups (permanent clearance).
  \item \textbf{Topology} $\mathcal{T}$: for each group, a connected graph
        (typically a tree), together with a feasible coarse polyline or grid
        embedding that initializes the refine.
  \item \textbf{Usages} $\mathcal{U}$: a set of directed terminal--terminal
        flows that must be considered when scoring fairness
        (e.g.\ $A\!\to\!C$, $B\!\to\!C$, $A\!\to\!B$).
  \item \textbf{Radii and ports}: tube radii along the layout, fixed terminal
        positions, and optional end-tangent / line-axis clamps so that a port
        sits on a prescribed axis.
\end{itemize}

The decision variables $X$ parameterize centerlines (and, implicitly,
tubular solids of the given radii).
The output must keep every terminal connected as in $\mathcal{T}$: each declared
usage remains a continuous path between its ends.
Branches need not meet at a single geometric point; Sec.~\ref{sec:problem:junction}
explains when a local split near a junction is allowed.

\subsection{Objective sketch}
\label{sec:problem:objective}

Given topology $\mathcal{T}$ and a feasible coarse embedding, refine the
geometric parameters $X$ so that a weighted soft cost over declared usages is
minimized, subject to hard geometric constraints:
\begin{equation}
\label{eq:stage1}
\min_{X}
\;
J_{\mathrm{path}}(X)
+
J_{\mathrm{clear}}(X)
+
J_{\mathrm{union}}(X)
+
J_{\mathrm{space}}(X)
\quad
\text{s.t.\ ports, }\mathcal{T}\text{, interface }C^{0}/C^{1}
\text{ by construction.}
\end{equation}
Table~\ref{tab:soft-hard} summarizes the roles of the terms.
Full discrete definitions appear in Sec.~\ref{sec:costs}; here we only fix the
semantics needed for the problem statement.

\begin{table}[H]
  \centering
  \caption{Soft versus hard elements in the geometric refine.}
  \label{tab:soft-hard}
  \begin{tabular}{@{}p{0.22\linewidth}p{0.48\linewidth}p{0.22\linewidth}@{}}
    \toprule
    Symbol & Role & Soft / hard \\
    \midrule
    $X$
      & Centerline parameters (shared trunk + private junction controls)
      & decision \\
    $L,K$ in $J_{\mathrm{path}}$
      & Chord energy (each shared piece once) + bend (segment interiors once;
        usage joints once)
      & soft \\
    $J_{\mathrm{clear}}$
      & Obstacle and cross-group tube separation
      & soft in loop; \textbf{hard} at accept \\
    $J_{\mathrm{union}}$
      & Junction-bundle compactness
      & soft \\
    $J_{\mathrm{space}}$
      & Keep sample gaps comparable under parameter sampling
      & soft \\
    Topology / ports / interface
      & Connectivity, fixed ends, $C^{0}/C^{1}$ clamps
      & \textbf{hard} (by construction) \\
    \bottomrule
  \end{tabular}
\end{table}

\subsection{Difference from classical path smoothing}
\label{sec:problem:diff}

\begin{table}[H]
  \centering
  \caption{Classical path smoothing versus the pipe refine studied here.}
  \label{tab:vs-smooth}
  \begin{tabular}{@{}p{0.22\linewidth}p{0.34\linewidth}p{0.34\linewidth}@{}}
    \toprule
    Aspect & Normal path smoothing & This problem \\
    \midrule
    Occupancy
      & Pass-and-release (time / moving agent)
      & Permanent tubes; clearance is static geometry \\
    Structure
      & Single $s$--$t$ line
      & Tree / branched manifold; degree $\ge 3$ joins \\
    Cost scope
      & One path's length / curvature
      & Usage-assembled paths over the branched manifold \\
    Domain
      & Often open or mild
      & Tight $\Omega$ may force large $\kappa$ (soft $K$) \\
    Ends
      & Fixed $s,t$ (pose optional)
      & Fixed ports plus line-axis alignment \\
    \bottomrule
  \end{tabular}
\end{table}

\subsection{Junction conflict: shared node versus local split}
\label{sec:problem:junction}

Usages matter because shared trunk nodes couple directions.
Optimizing only the shared polyline can make $A\!\to\!C$ and $B\!\to\!C$ smooth
while $A\!\to\!B$ remains a hairpin at the branch.
The declared set $\mathcal{U}$ therefore decides which end-to-end paths must stay
fair, and where junctions need private degrees of freedom.

Consider a Y-junction with fixed topology and usages
$\mathcal{U}=\{A\!\to\!C,\, B\!\to\!C,\, A\!\to\!B\}$:
\[
A\text{---}J\text{---}C,\qquad B\text{---}J.
\]

\begin{figure}[H]
  \centering
  \includegraphics[width=\linewidth]{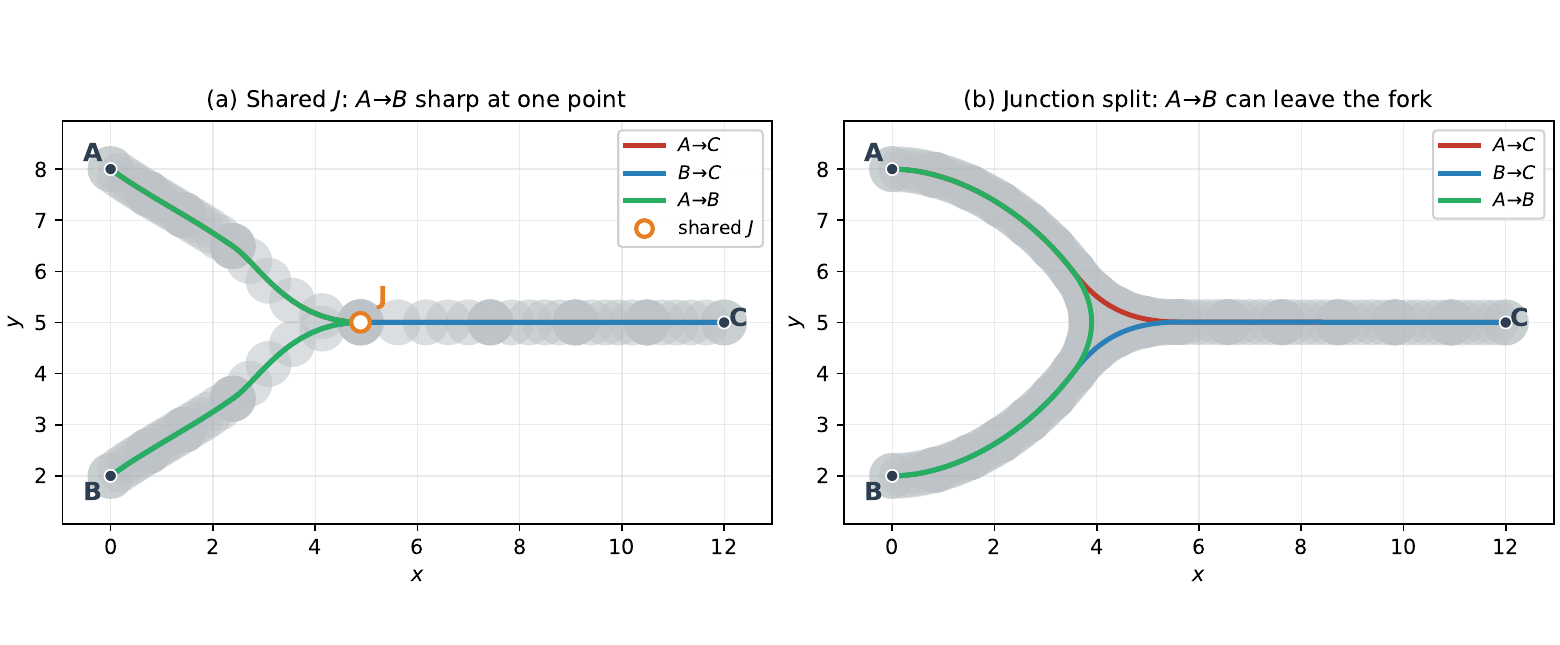}
  \caption{Planar Y under the same usages
  $\mathcal{U}=\{A\!\to\!C,\,B\!\to\!C,\,A\!\to\!B\}$ (top view; gray disks =
  tube radius).
  (a)~Forced shared connect point: the three usage centerlines meet at one
  node $J$ (orange); refine can fair $A\!\to\!C$ and $B\!\to\!C$, but
  $A\!\to\!B$ (green) stays sharp at $J$.
  (b)~Hybrid trunk-share / junction-split: $A\!\to\!B$ can leave the fork
  while the trunk toward $C$ remains shared.}
  \label{fig:y-shape}
\end{figure}

\paragraph{Shared junction.}
All three branches meet at one point $J$.
Fair geometry for $A\!\to\!C$ and $B\!\to\!C$ forces a sharp turn on
$A\!\to\!B$ at $J$ (Fig.~\ref{fig:y-shape}a).

\paragraph{Locally split junction.}
Private junction degrees of freedom let $A\!\to\!B$ take a dedicated arc while
$A\!\to\!C$ and $B\!\to\!C$ keep a shared trunk (Fig.~\ref{fig:y-shape}b).
A void appears near the join; connectivity of $\mathcal{T}$ is preserved, but
the embedding need not keep a common junction point.

\begin{table}[H]
  \centering
  \caption{Shared junction versus locally split junction.}
  \label{tab:junction-regimes}
  \begin{tabular}{@{}lll@{}}
    \toprule
    & Shared junction & Locally split junction \\
    \midrule
    Geometry at join
      & one shared point $J$
      & trunks + dedicated $A\!\to\!B$ arc + void \\
    $A\!\to\!C$, $B\!\to\!C$
      & arc + straight
      & same as shared case \\
    $A\!\to\!B$
      & sharp at $J$
      & dedicated arc; fair \\
    $\mathcal{T}$
      & preserved
      & preserved (connectivity, not a common join point) \\
    \bottomrule
  \end{tabular}
\end{table}

The design implication is that the refine must allow private degrees of freedom
near junctions (local split) while sharing trunk controls where usages truly share
physical pipe.
That hybrid structure---trunk-share + junction-split---is the method focus of
this report.
The core gap relative to single-curve fairing is therefore
\textbf{topology-preserving, multi-usage geometric refinement under permanent
tubular occupancy}, not merely denser spline fitting.

\subsection{Place in the pipeline}
\label{sec:problem:pipeline}

A typical valve-channel workflow is discrete multi-group routing, then
centerline refine, then CAD solids, then fluid PDE / shape polish.
This report covers only the geometric refine.

That ordering matters for junctions.
Downstream shape or adjoint PDE optimization usually deforms a
\emph{fixed} channel layout: it can move walls or a shared centerline, but it
does not introduce a new split curve or triangular void at a Y-join the way
a locally split junction does.
That local split is therefore a decision of the geometric refine, not of
later physics polish.
We still preserve the connectivity of $\mathcal{T}$ (no discrete rewiring);
what changes is the geometric arrangement near degree-$\ge 3$ nodes.
Pressure drop and streamline quality remain for subsequent physics-based
optimization, once the centerline graph---including any local splits---is
fixed.

%% file: sec_related.tex
% !TEX root = main.tex
% =============================================================================
% Related work (compact for technical report)
% =============================================================================

\section{Related Work}
\label{sec:related}

\paragraph{Discrete pipe routing.}
Belov et al.~\cite{belov2020pr} adapt MAPF-style search to 3D pipe routing;
surveys of plant-scale automatic routing emphasize sequential CAD and
MIP/sampling pipelines~\cite{blokland2023pipe}.
Our companion report~\cite{deng2025pipe_design} treats \emph{multi-group}
routing under permanent occupancy and supplies the coarse layouts we refine.
That work stops at discrete embeddings; continuous centerline cleanup is left
open.

\paragraph{Robotic path smoothing.}
CHOMP, STOMP, TEB, and TrajOpt~\cite{zucker2013chomp,kalakrishnan2011stomp,roesmann2013teb,schulman2014trajopt}
refine a single temporary trajectory under smoothness and collision costs.
We inherit the gradient-centric idea, but a one-path objective does not encode
shared trunks, permanent cross-group tubes, or the multi-usage junction conflict
of Sec.~\ref{sec:problem:junction}.

\paragraph{Fairing, CAD, and physics polish.}
Elastic-rod / curve fairing~\cite{moreton1992functional,bergou2008discrete}
and clothoid primitives~\cite{fraichard2004from} motivate soft bend energies on
one open curve; sparse circular-arc encodings are a poor fit for dense
grid-routed manifolds.
Parametric Hermite / NURBS centerlines are standard in CAD~\cite{piegl1997nurbs},
yet manual fillets do not resolve multi-usage joins under hard clearance.
Adjoint CFD / shape optimization sits \emph{downstream}: it typically deforms a
fixed channel layout and does not create a locally split junction
(Sec.~\ref{sec:problem:pipeline}).
Our contribution is the combination---branched, usage-coupled refine under
permanent tubular occupancy---not any single fairing or collision ingredient.

%% file: sec_method.tex
% !TEX root = main.tex
% =============================================================================
% Method: hybrid trunk-share / junction-split + Hermite
% Cost formulas → sec_costs (next).
% =============================================================================

\section{Method}
\label{sec:method}

This section fixes the parameterization and the optimize loop; cost formulas
are in Sec.~\ref{sec:costs}.

\subsection{Pipeline overview}
\label{sec:method:pipeline}

For each group independently (cross-group coupling enters only through
clearance):

\begin{enumerate}
  \item \textbf{Tree.} Deduplicate the coarse grid / polyline embedding into an
        undirected tree $\mathcal{T}$ (or forest): nodes are terminals and
        branch points; edges carry the coarse polylines and radii.
  \item \textbf{Junction partition.} Grow junction complexes around
        degree-$\ge 3$ nodes; label remaining edges as trunks
        (Sec.~\ref{sec:method:partition}).
  \item \textbf{Hermite graph.} Instantiate cubic Hermite segments on labeled
        edges. Trunk knots are shared; junction knots are private per labeled
        junction edge; ports and trunk--junction interfaces are clamped for
        $C^{0}/C^{1}$ by construction (Sec.~\ref{sec:method:hermite}).
  \item \textbf{Assemble \& optimize.} Sample usages, minimize soft $J$ by
        gradient steps, accept under hard clearance (Alg.~\ref{alg:refine}).
\end{enumerate}

Groups couple only through $J_{\mathrm{clear}}$ and the final accept test.
Topology $\mathcal{T}$ and $\mathcal{U}$ are never rewritten.

\subsection{Junction partition: where DOF are private}
\label{sec:method:partition}

The hybrid rule is: \emph{share controls wherever usages share physical pipe;
add private controls only where multi-usage curvature conflicts}.

\paragraph{Grow complexes.}
Seed every tree node of degree $\ge 3$.
Grow a zone around each seed by tree-path length with threshold
$\tau=\max(\tau_0,\,\alpha\,r_{\mathrm{local}},\,\beta\,R_{\mathrm{bend}})$.
After the zone grow, any connected trunk component whose total length is shorter
than a threshold $\tau_{\mathrm{trunk}}$ is absorbed into junction (so nearby
seeds separated by a short trunk become one complex $Z$).
Edges inside $Z$ are labeled \texttt{junction}; edges outside all complexes are
\texttt{trunk}.

\paragraph{Variable layout.}
Table~\ref{tab:regions} summarizes roles after the label pass.
When building the Hermite graph we apply one privatization rule:
\emph{ports and complex interfaces keep a single global knot; every other
sample on a labeled edge is private to that edge}, so two junction edges that
meet at the same coarse branch point do not share interior $(p,v)$.
Usages do not each own a full duplicate spline; they inherit geometry by
walking the labeled edges on their tree path.
Junction chains stay short, so new freedom appears only near the conflict in
Fig.~\ref{fig:y-shape}.

\begin{table}[H]
  \centering
  \caption{Region roles after junction partition.}
  \label{tab:regions}
  \begin{tabular}{@{}p{0.16\linewidth}p{0.28\linewidth}p{0.24\linewidth}p{0.24\linewidth}@{}}
    \toprule
    Region & Geometry & Variables & Continuity \\
    \midrule
    Trunk
      & Edges outside all complexes
      & Shared Hermite knots $X_{\mathrm{trunk}}$
      & Intrinsic (one curve) \\
    Junction
      & Interior of complex $Z$
      & Private knots \emph{per labeled junction edge}
      & Meets trunk at interface \\
    Interface
      & Complex boundary
      & Shared endpoints (+ clamped tangents)
      & Hard $C^{0}$; hard $C^{1}$ by construction \\
    Ports
      & Group terminals
      & Fixed position / axis tangent
      & Hard clamps \\
    \bottomrule
  \end{tabular}
\end{table}

\paragraph{Y example (how a local split appears).}
On $A$--$J$--$C$ with $B$--$J$ and
$\mathcal{U}=\{A\!\to\!C,\,B\!\to\!C,\,A\!\to\!B\}$, the coarse embedding places
all three branches at one point $J$.
After partition, each usage is a short alternating chain of
\emph{shared trunks} and \emph{private junction edges}:
\begin{align*}
A\!\to\!C &:\;
  A_{\mathrm{trunk}}
  \;\to\;
  AC_{\mathrm{junc}}
  \;\to\;
  C_{\mathrm{trunk}},
\\
B\!\to\!C &:\;
  B_{\mathrm{trunk}}
  \;\to\;
  BC_{\mathrm{junc}}
  \;\to\;
  C_{\mathrm{trunk}},
\\
A\!\to\!B &:\;
  A_{\mathrm{trunk}}
  \;\to\;
  AB_{\mathrm{junc}}
  \;\to\;
  B_{\mathrm{trunk}}.
\end{align*}
Here $A_{\mathrm{trunk}}$, $B_{\mathrm{trunk}}$, and $C_{\mathrm{trunk}}$ are
shared Hermite geometry (every usage that enters that arm reads the same knots).
The three junction edges $AC_{\mathrm{junc}}$, $BC_{\mathrm{junc}}$, and
$AB_{\mathrm{junc}}$ are independent: each has its own interior $(p,v)$, even
though they all start at the same coarse coordinates near $J$.
Thus $A\!\to\!C$ and $B\!\to\!C$ couple through $C_{\mathrm{trunk}}$, while
$A\!\to\!B$ can fair through $AB_{\mathrm{junc}}$ without forcing a hairpin at a
single shared point.
Initialization copies the shared-junction layout, so the private junction knots
begin coincident; during optimization they may separate and open the void in
Fig.~\ref{fig:y-shape}b---a continuous motion of already-private DOF, not a
discrete mode switch.

\subsection{Cubic Hermite representation}
\label{sec:method:hermite}

Each labeled edge is a chain of cubic Hermite segments.
A segment with ends $(p_0,v_0)$ and $(p_1,v_1)$ is, for $t\in[0,1]$,
\begin{equation}
\label{eq:hermite}
\gamma(t)
=
H_{00}(t)\,p_0
+ H_{10}(t)\,v_0
+ H_{01}(t)\,p_1
+ H_{11}(t)\,v_1,
\end{equation}
with the standard Hermite basis
\begin{align*}
H_{00}(t) &= 2t^{3}-3t^{2}+1, &
H_{10}(t) &= t^{3}-2t^{2}+t, \\
H_{01}(t) &= -2t^{3}+3t^{2}, &
H_{11}(t) &= t^{3}-t^{2}.
\end{align*}
Thus $\gamma(0)=p_0$, $\gamma(1)=p_1$, $\gamma'(0)=v_0$, $\gamma'(1)=v_1$.
Shared nodes enforce hard $C^{0}$ and $C^{1}$: every incident segment reads the
same $(p,v)$.

\paragraph{Clamps.}
Port nodes fix $p$ and align $v$ with the prescribed unit vector.
Interface nodes between trunk and junction share $p$ (and, by construction,
compatible $v$) so that usage walks stay $G^{1}$ across the complex boundary.
Only movable trunk knots and private junction knots enter the decision
vector $x$.

\subsection{Assemble and optimize}
\label{sec:method:opt}

Alg.~\ref{alg:refine} minimizes the soft objective of Sec.~\ref{sec:costs} by
first-order steps on $\nabla_x J$ and accepts a design only under hard obstacle /
cross-group clearance.
Cost formulas are in Sec.~\ref{sec:costs}; which first-order variant we use in
practice is an implementation detail reported with the experiments.

\begin{algorithm}[H]
\caption{Usage assembly and gradient refine (per multi-group instance)}
\label{alg:refine}
\begin{algorithmic}[1]
\Require Hermite graphs $\{G_g\}$, usages $\mathcal{U}$, weights, steps $T$, refresh period $k$
\State $x \leftarrow$ concatenate free node positions and tangents $(p,v)$ across groups
\For{$t = 1,\ldots,T$}
  \State write $x$ into the Hermite graphs
  \If{$t=1$ or $(t-1) \bmod k = 0$}
    \State find candidate colliding pairs by proximity (broadphase):
    \Statex \quad $\mathcal{N}_{\mathrm{obs}}$: tube samples vs obstacle primitives
    \Statex \quad $\mathcal{N}_{\mathrm{grp}}$: tube samples of group $g$ vs group $g'$
    \Statex \quad $\mathcal{N}_{\mathrm{union}}$: nearby junction samples within a complex
  \EndIf
  \For{each usage $u \in \mathcal{U}$}
    \State follow $u$'s path on $\mathcal{T}$ (shared trunks + private junctions)
    \State sample Hermite segments along that path $\to \gamma_u,\, r_u$
  \EndFor
  \State $J_{\mathrm{path}} \leftarrow$ length / curvature on $\{\gamma_u,r_u\}$
        \Comment{Sec.~\ref{sec:costs}}
  \State $J_{\mathrm{space}} \leftarrow$ spacing penalty on chord gaps
  \State $J_{\mathrm{clear}} \leftarrow$ soft gaps on pairs in
        $\mathcal{N}_{\mathrm{obs}}\cup\mathcal{N}_{\mathrm{grp}}$
  \State $J_{\mathrm{union}} \leftarrow$ soft compactness on pairs in
        $\mathcal{N}_{\mathrm{union}}$
  \State $J \leftarrow J_{\mathrm{path}}+J_{\mathrm{space}}+J_{\mathrm{clear}}+J_{\mathrm{union}}$
  \State differentiate $J$ w.r.t.\ $x$ through the sampling map (reverse-mode AD)
  \State $x \leftarrow$ gradient-based update using $\nabla_x J$
\EndFor
\State write $x$ back; hard-check obstacle / cross-group clearance (no soft margin)
\State \textbf{accept} iff hard hit count $= 0$; else reject / raise clearance weights
\Ensure accepted Hermite graphs, or failure
\end{algorithmic}
\end{algorithm}

%% file: sec_costs.tex
% !TEX root = main.tex
% =============================================================================
% Cost functions
% =============================================================================

\section{Cost Functions}
\label{sec:costs}

Alg.~\ref{alg:refine} minimizes a scalar soft objective
\begin{equation}
\label{eq:J}
J(x)
=
J_{\mathrm{path}}(x)
+
J_{\mathrm{clear}}(x)
+
J_{\mathrm{union}}(x)
+
J_{\mathrm{space}}(x).
\end{equation}
Terms below are evaluated on Hermite samples (and proximity pairs from
Alg.~\ref{alg:refine}).

\subsection{Path costs}
\label{sec:costs:path}

We set
\begin{equation}
J_{\mathrm{path}}
=
\alpha\,L
+
\beta\,K.
\end{equation}
Let $\mathcal{E}$ be the set of Hermite segments (trunk and junction) and
$\mathcal{J}$ the set of unordered segment--segment joints discovered by walking
usages.

\paragraph{Aggregation (not a naive usage sum).}
Chord energy charges each physical segment once,
\begin{equation}
L
=
\sum_{e\in\mathcal{E}}
L_e,
\end{equation}
so a shared trunk is not multiplied by how many usages traverse it.
Curvature sums segment interiors once and every distinct usage joint once,
\begin{equation}
K
=
\sum_{e\in\mathcal{E}}
K_e^{\mathrm{int}}
+
\sum_{\{e,e'\}\in\mathcal{J}}
K_{e,e'}^{\mathrm{joint}}.
\end{equation}
Walking $\mathcal{U}$ is still required for $K$: the hairpin at a shared $Y$
lives at \emph{which} segments meet, even when trunk interiors were already
scored.
A naive $\sum_u L_u$ (or $\sum_u K_u$ over full usage polylines) would
over-count shared material and over-pull trunks.

\paragraph{Chord energy.}
On a sample chain $\{x_i\}$, we use squared chords rather than
$\sum_i\|x_{i+1}-x_i\|_2$:
\begin{equation}
L_e
=
\sum_{i}
\|x_{i+1}-x_i\|_2^{2}
\end{equation}
(with optional sample weights $\bar w_i=\tfrac12(w_i+w_{i+1})$).
Squaring penalizes uneven gap distributions more strongly than raw chord length
(one long chord costs more than several short ones of equal total length).

\paragraph{Radius-aware bend.}
The same centerline turn is worse for a fatter tube, so we penalize a
dimensionless bend related to $\kappa r$, not bare $\kappa$.
With unit chords $t_i$ and turning angle
\begin{equation}
\theta_i
=
\mathrm{atan2}\big(\|t_{i-1}\times t_i\|_2,\; t_{i-1}\cdot t_i\big)
\in[0,\pi]
\end{equation}
(preferred over $\arccos(t_{i-1}\cdot t_i)$, which is ill-behaved as
$\theta\!\to\!0$), the bend energy on a chain is
\begin{equation}
\label{eq:Kbend}
\sum_{i}
\frac{(\theta_i\, r_i)^{2}}{\Delta\ell_i+\varepsilon}
\approx
\int (\kappa r)^{2}\,\mathrm{d}s,
\end{equation}
with chord lengths $\Delta s_i=\|x_{i+1}-x_i\|_2$ and
$\Delta\ell_i=\tfrac12(\Delta s_{i-1}+\Delta s_i)$.
The same kernel is applied on segment interiors ($K_e^{\mathrm{int}}$) and on
three-point windows at usage joints ($K_{e,e'}^{\mathrm{joint}}$).

\subsection{Clearance}
\label{sec:costs:clear}

For two capsule / ball witnesses with centers $a,b$ and radii $r_a,r_b$, let
$d_{\mathrm{req}}=r_a+r_b+\delta_{\mathrm{margin}}$ and
\begin{equation}
\phi(a,b;r_a,r_b)
=
\Big(
\big[
d_{\mathrm{req}}-\|a-b\|_2
\big]_+^{\sigma}
\Big)^{2},
\end{equation}
where $[\cdot]_+^\sigma$ is a softplus hinge.
Then
\begin{equation}
J_{\mathrm{clear}}
=
J_{\mathrm{obs}}
+
J_{\mathrm{grp}},
\end{equation}
with $J_{\mathrm{obs}}$ summing $\phi$ over tube--obstacle pairs in
$\mathcal{N}_{\mathrm{obs}}$ and $J_{\mathrm{grp}}$ summing $\phi$ over
cross-group pairs in $\mathcal{N}_{\mathrm{grp}}$
(Alg.~\ref{alg:refine}).
Weights $\mu_{\mathrm{obs}}$ and $\mu_{\mathrm{grp}}$ scale the two sums.

\paragraph{Hard accept.}
After the loop we recompute residuals on obstacle / cross-group pairs
\emph{without} the soft margin used inside $J_{\mathrm{clear}}$ and accept only
if the hard hit count is zero (Alg.~\ref{alg:refine}).
Capsule proxies are incomplete relative to exact swept solids, but sufficient
for the geometric refine before CAD.

\subsection{Junction union}
\label{sec:costs:union}

Private junction patches are under-determined by $L$ and $K$ alone.
Inside each complex $Z$, let $\gamma_u^{Z}$ be the samples of usage $u$ on
junction edges of $Z$, and let $\mathcal{U}_Z$ be the usages that enter $Z$.
For a point $x$ with radius $r_x$ and a curve $\gamma$ with nearest-sample
radius $r_\star$, set $r_{\Sigma}=r_x+r_\star$ and
\begin{equation}
\psi(d;r_{\Sigma})
=
\begin{cases}
\displaystyle\bigl(d/r_{\Sigma}\bigr)^{2}
  & d < \alpha\, r_{\Sigma}, \\[0.4em]
0 & \text{otherwise,}
\end{cases}
\end{equation}
with cutoff factor $\alpha\sim O(1)$.
The pairwise Chamfer
\begin{equation}
\mathrm{Ch}(\gamma_u^{Z},\gamma_v^{Z})
=
\sum_{x\in\gamma_u^{Z}}
\psi\big(d(x,\gamma_v^{Z});r_{\Sigma}\big)\,\Delta s(x)
\;+\;
\sum_{y\in\gamma_v^{Z}}
\psi\big(d(y,\gamma_u^{Z});r_{\Sigma}\big)\,\Delta s(y)
\end{equation}
then defines
\begin{equation}
J_{\mathrm{union}}
=
\lambda
\sum_{Z}
\sum_{\{u,v\}\subset\mathcal{U}_Z}
\mathrm{Ch}(\gamma_u^{Z},\gamma_v^{Z}).
\end{equation}
While usages remain within about one radius sum, $\psi$ pulls them into a
compact bundle; once they separate beyond the cutoff (local split,
Fig.~\ref{fig:y-shape}b), the force releases.
Radius normalization keeps $\lambda$ comparable across tube sizes.

\subsection{Spacing regularizer}
\label{sec:costs:space}

Hermite segments are sampled in the parameter $t\in[0,1]$, not in arc length.
Without a spacing term, the optimizer can stretch $\|\gamma'(t)\|$ so that large
gaps open between samples: soft clearance and bend then see only a sparse,
biased subset of the curve.
We therefore use
\begin{equation}
J_{\mathrm{space}}
=
\eta
\sum_{i}
\big(\Delta s_i-\overline{\Delta s}\big)^{2},
\end{equation}
where $\overline{\Delta s}$ is the mean chord on the chain, to keep sample gaps
comparable and reduce this parameterization artifact.

%% file: sec_experiments.tex
% !TEX root = main.tex
% =============================================================================
% Experiments
% =============================================================================

\section{Experiments}
\label{sec:experiments}

We evaluate the hybrid refine on a synthetic multi-group suite with capsule
clearance.
The goal is to check that the method is usable at scale: hard clearance accept,
dimensionless bend $\kappa r\lesssim 1$, and a clearly decreasing soft objective.

\subsection{Setup}
\label{sec:exp:setup}

\paragraph{Instances.}
We use $N{=}50$ instances from \texttt{hard\_r2}: each is a $48^3$ grid layout
from the companion discrete router~\cite{deng2025pipe_design}
(\textsc{PlaneSlice} corridors + \textsc{PillarField} obstacles), with
\textbf{8 groups}, \textbf{8 terminals per group}, and uniform tube radius
$r{=}2$.
Coarse paths and declared usages are fixed inputs; the refine does not replan
topology.

\paragraph{Protocol.}
All runs use capsule clearance, Chamfer junction union, and the costs of
Sec.~\ref{sec:costs}.
Free Hermite DOF are updated for a budget of $1500$ first-order steps
(Alg.~\ref{alg:refine}) with step size $0.01$, soft clearance margin $0.15$,
and proximity-pair refresh every $10$ steps.
Weights are
$\alpha{=}1$, $\beta{=}1$, $\mu_{\mathrm{obs}}{=}1$, $\mu_{\mathrm{grp}}{=}1$,
$\lambda{=}0.1$, $\eta{=}10$.

\paragraph{Metrics.}
\begin{itemize}
  \item \textbf{Success:} fraction of instances with hard hit count
        $n_{\mathrm{hit}}=0$ (obstacle and cross-group; no soft margin).
  \item \textbf{Path:} aggregate true polyline length $L$, max dimensionless
        bend $\kappa r_{\max}$, and voxel tube volume $V$ (accepted runs).
  \item \textbf{Solve:} wall time, final soft objective $J_T$, and relative drop
        $(J_0-J_T)/J_0$.
\end{itemize}

\subsection{Results}
\label{sec:exp:results}

Table~\ref{tab:exp-main} reports the batch.
All $50$ instances are accepted.
Mean $\kappa r_{\max}\approx 0.97$ stays below one, and the soft objective drops
by about $95\%$ on average
($J_0\!\sim\!6.55\times 10^{3}\!\to\!J_T\!\sim\!3.60\times 10^{2}$).
Mean wall time is about $160\,\mathrm{s}$ per instance under this protocol.

\begin{table}[H]
  \centering
  \caption{Hard\_r2 refine with capsule clearance ($N{=}50$).
  Means over all runs (all accepted).}
  \label{tab:exp-main}
  \begin{tabular}{@{}ccccccc@{}}
    \toprule
    Success & Time (s) & $J_T$ & Rel.\ drop & $L$ & $\kappa r_{\max}$ & $V$ \\
    \midrule
    $50/50$ & $160$ & $360$ & $0.95$ & $4298$ & $0.97$ & $1.91{\times}10^{4}$ \\
    \bottomrule
  \end{tabular}
\end{table}

\begin{figure}[H]
  \centering
  \begin{minipage}[t]{0.495\linewidth}
    \centering
    \includegraphics[width=\linewidth]{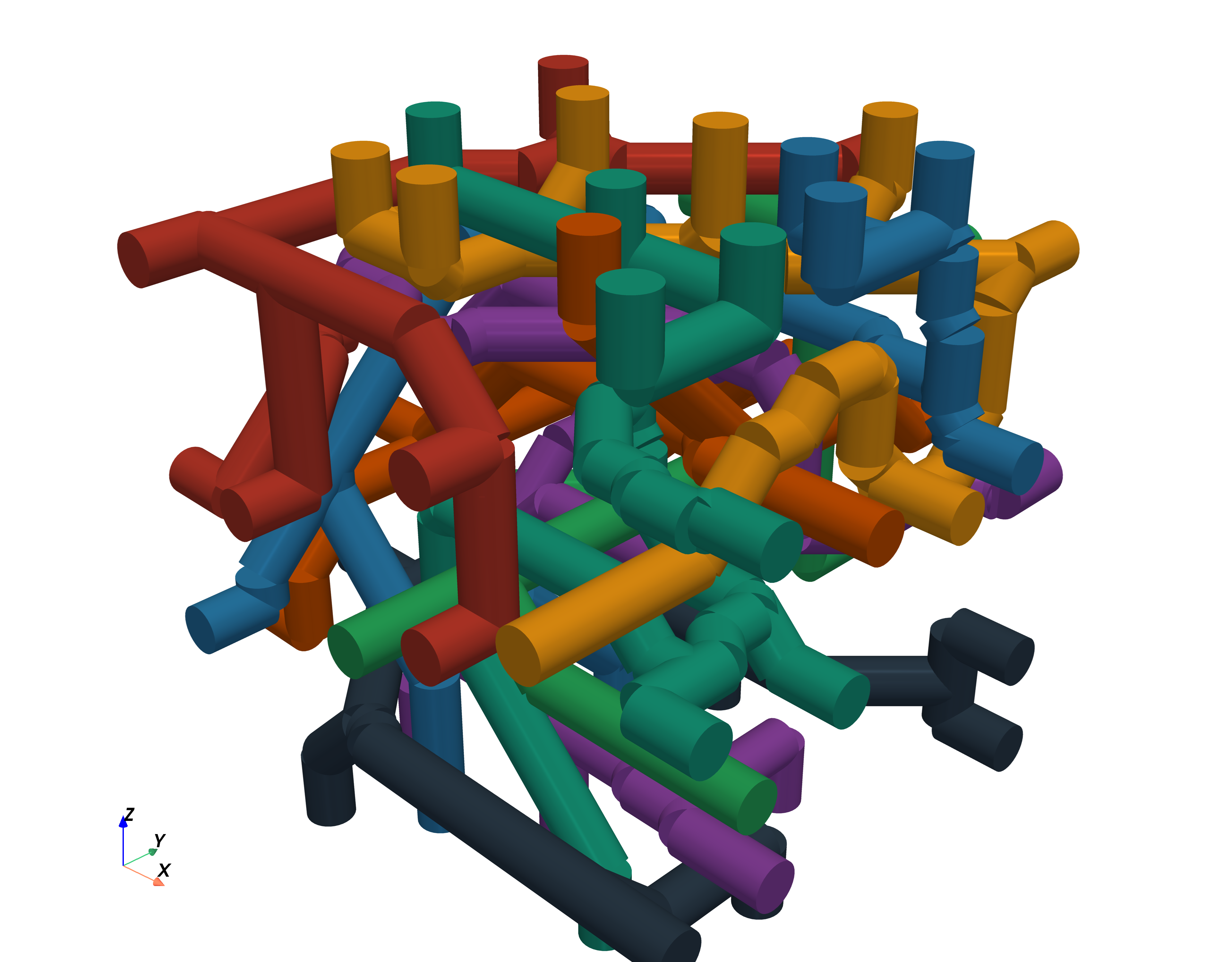}\\[0.35em]
    {\small Before (coarse)}
  \end{minipage}\hfill
  \begin{minipage}[t]{0.495\linewidth}
    \centering
    \includegraphics[width=\linewidth]{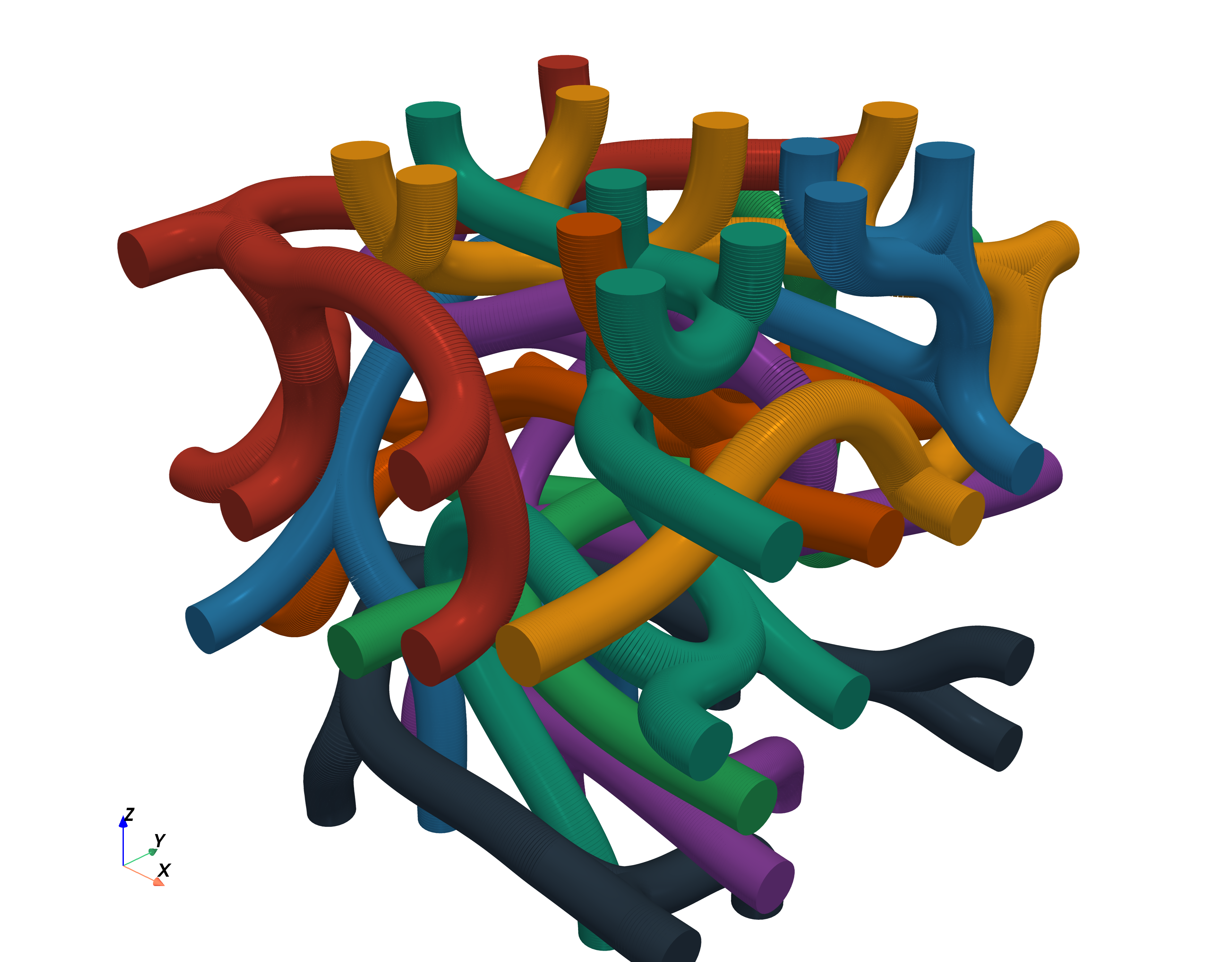}\\[0.35em]
    {\small After (refined)}
  \end{minipage}
  \caption{All groups of instance~0 from \texttt{hard\_r2}: coarse discrete
  paths (left) vs.\ refined Hermite tubes (right). Colors mark groups;
  obstacles omitted.}
  \label{fig:exp-all}
\end{figure}

\begin{figure}[H]
  \centering
  \begin{minipage}[t]{0.495\linewidth}
    \centering
    \includegraphics[width=\linewidth]{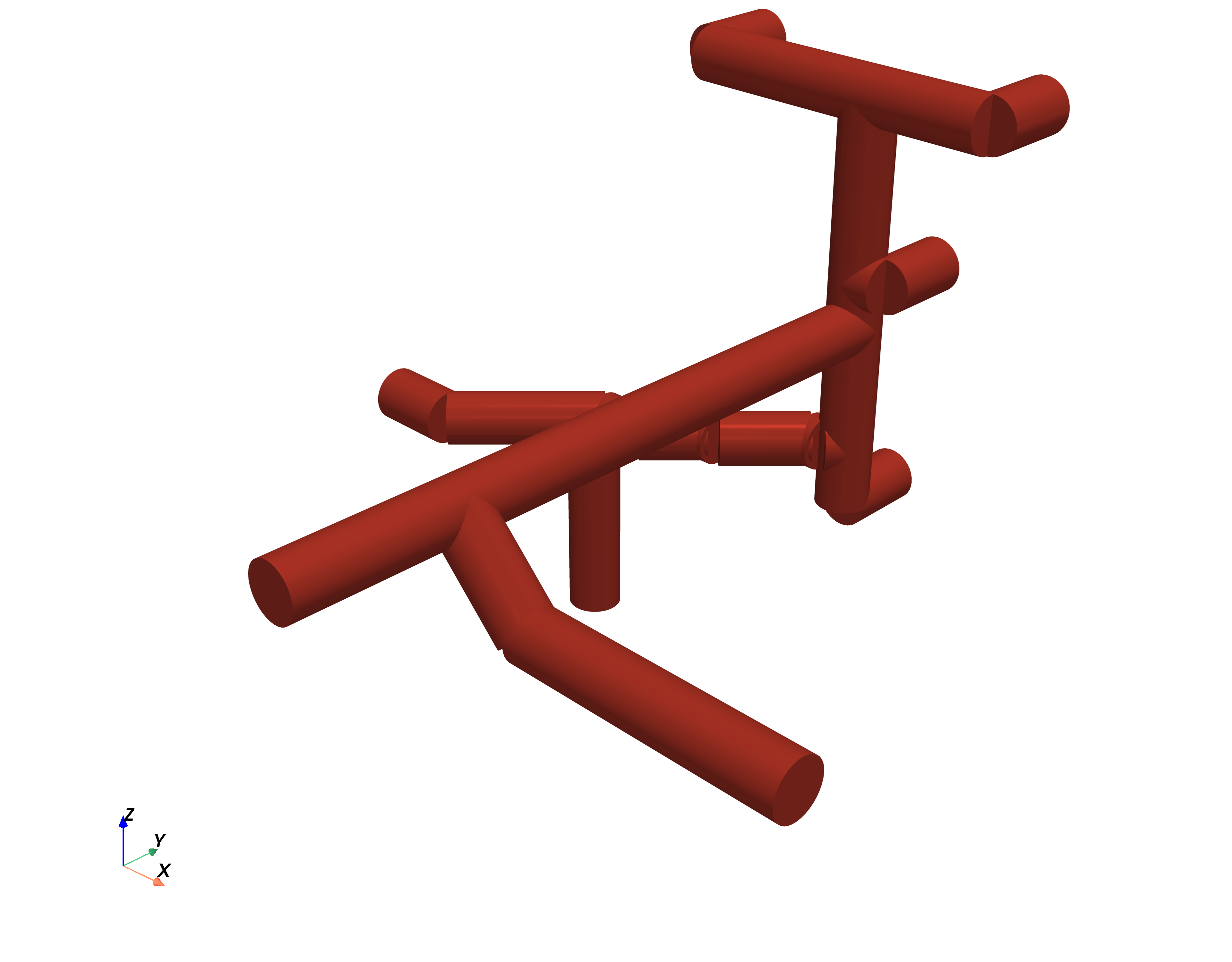}\\[0.35em]
    {\small Before (coarse)}
  \end{minipage}\hfill
  \begin{minipage}[t]{0.495\linewidth}
    \centering
    \includegraphics[width=\linewidth]{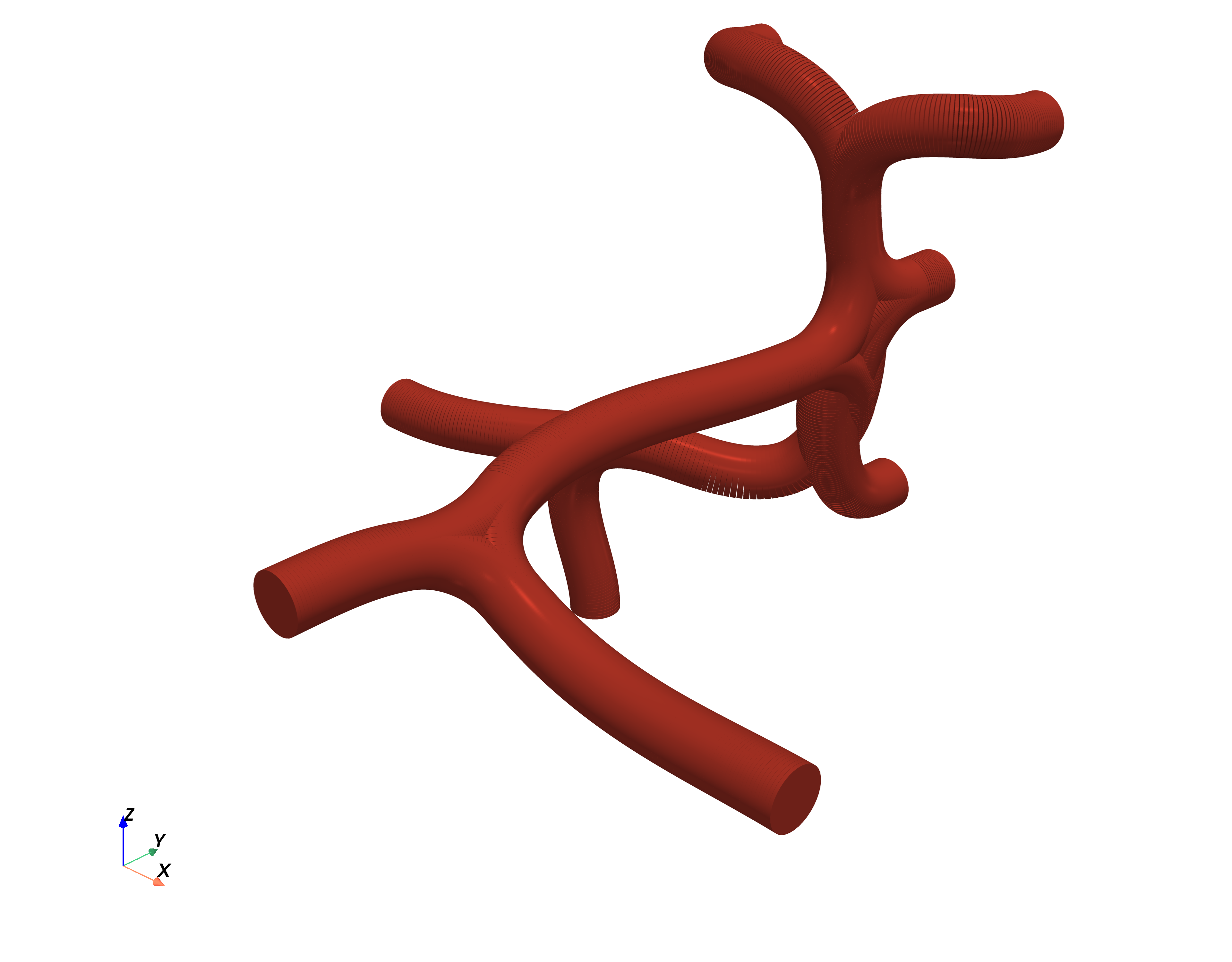}\\[0.35em]
    {\small After (refined)}
  \end{minipage}
  \caption{Single group ($g_2$) of the same instance, obstacles omitted:
  coarse discrete paths (left) vs.\ refined Hermite tubes (right).}
  \label{fig:exp-pair}
\end{figure}

\begin{figure}[H]
  \centering
  \includegraphics[width=0.72\linewidth]{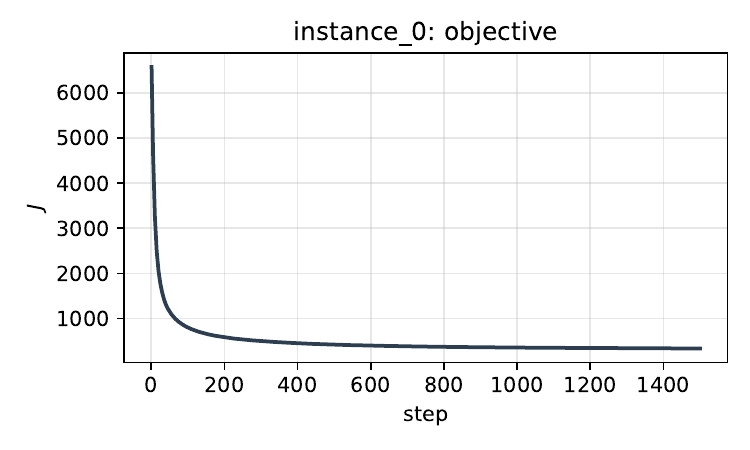}
  \caption{Soft objective $J$ vs.\ iteration for instance~0
  (\texttt{hard\_r2\_w1} / Adam / capsule).}
  \label{fig:exp-J}
\end{figure}

Fig.~\ref{fig:exp-all} shows the full multi-group packing of instance~0
before and after refine (colors mark groups; obstacles omitted for clarity).
Fig.~\ref{fig:exp-pair} isolates one group so that grid corners vs.\ fair
Hermite tubes are easier to read.
Fig.~\ref{fig:exp-J} plots the soft objective for that run; $J$ drops sharply
in the first few hundred steps and then plateaus.

%% file: sec_conclusion.tex
% !TEX root = main.tex
% =============================================================================
% Conclusion (compact; no separate Discussion)
% =============================================================================

\section{Conclusion}
\label{sec:conclusion}

We formalized topology-preserving, multi-usage pipe centerline refinement under
permanent tubular occupancy, and instantiated it with hybrid trunk-share /
junction-split Hermite variables, soft path / clearance / union / spacing
costs, and hard clearance at accept.
On $50$ hard multi-group synthetic instances with capsule clearance, the refine
reaches full hard-accept success, $\kappa r_{\max}\lesssim 1$, and about a
$95\%$ relative drop in the soft objective.

The objective here is geometric: $K$ is a bend proxy, not pressure drop, and
capsule clearance is a fast narrowphase rather than exact CAD interference.
Fluid PDE polish and automatic weight schedules are left to later work; this
report only claims a usable geometric stage between discrete multi-group
routing and subsequent solid / physics optimization.